\documentclass[a4paper,conference]{IEEEtran}

\usepackage{cite}

\usepackage[pdftex]{graphicx}
\usepackage{amsmath}
\usepackage{amssymb}
\usepackage{algorithmic}
\usepackage{array}
\usepackage{subfig}
\usepackage{url}            

\usepackage{fixltx2e}

\usepackage{stfloats}
\usepackage{url}

\begin{document}

\title{\emph{DeepScores} -- A Dataset for Segmentation, Detection and Classification of Tiny Objects}

\author{\IEEEauthorblockN{Lukas Tuggener}
\IEEEauthorblockA{ZHAW Datalab \& USI\\
tugg@zhaw.ch}
\and
\IEEEauthorblockN{Ismail Elezi}
\IEEEauthorblockA{University of Venice \& ZHAW \\
ismail.elezi@unive.it}
\and
\IEEEauthorblockN{J{\"u}rgen Schmidhuber}
\IEEEauthorblockA{IDSIA \& USI\\
juergen@idsia.ch}
\and
\IEEEauthorblockN{Marcello Pelillo}
\IEEEauthorblockA{University of Venice\\
pelillo@unive.it}
\and 
\IEEEauthorblockN{Thilo Stadelmann}
\IEEEauthorblockA{ZHAW Datalab\\
stdm@zhaw.ch}}

\maketitle

\begin{abstract}
We present the DeepScores dataset with the goal of advancing the state-of-the-art in small object recognition by placing the question of object recognition in the context of scene understanding. DeepScores contains high quality images of musical scores, partitioned into $300,000$ sheets of written music that contain symbols of different shapes and sizes. With close to a hundred million small objects, this makes our dataset not only unique, but also the largest public dataset. DeepScores comes with ground truth for object classification, detection and semantic segmentation. DeepScores thus poses a relevant challenge for computer vision in general, and optical music recognition (OMR) research in particular. We present a detailed statistical analysis of the dataset, comparing it with other computer vision datasets like PASCAL VOC, SUN, SVHN, ImageNet, MS-COCO, as well as with other OMR datasets. Finally, we provide baseline performances for object classification, intuition for the inherent difficulty that DeepScores poses to state-of-the-art object detectors like YOLO or R-CNN, and give pointers to future research based on this dataset.
\end{abstract}


\IEEEpeerreviewmaketitle

\section{Introduction}
Increased availability of data and computational power has often been followed by progress in computer vision and machine learning. The recent rise of deep learning in computer vision for instance has been promoted by the availability of large image datasets \cite{Deng2009} and increased computational power provided by GPUs \cite{gpu2004, Raina2009, ciresan2012cvpr}.

Optical music recognition (OMR) \cite{Rebelo2012} is a classical and challenging area of document recognition and computer vision that aims at converting scans of written music to machine-readable form, much like optical character recognition (OCR) \cite{mori1999} does for printed text. While results on simplified tasks show promising results \cite{vanderWel17,jorge2017}, there is yet no OMR solution that leverages the power of deep learning. We conjecture that this is caused in part by the lack of publicly available datasets of written music, big enough to train deep neural networks. The \emph{DeepScores} dataset has been collected with OMR in mind, but as well addresses important aspects of next generation computer vision research that pertain to the size and number of objects per image.

Although there is already a number of clean, large datasets available to the computer vision community \cite{Deng2009, Xiao2010, Everingham2010, Netzer2011, Lin2014}, those datasets are similar to each other in the sense that for each image there are a few large objects of interest. Object detection approaches that have shown state-of-the-art performance under these circumstances, such as Faster R-CNN \cite{Ren2014}, SSD \cite{Liu2016} and YOLO \cite{Redmon2016}, demonstrate very poor off-the-shelf performances when applied to environments with large input images containing multiple small objects (see Section \ref{sec_baselines}). 

Sheets of written music, on the other hand, usually have dozens to hundreds of small salient objects. The class distribution of musical symbols is strongly skewed and the symbols have a large variability in size. Additionally, the OMR problem is very different from modern OCR \cite{Goodfellow2013, Lee16}: while in classical OCR, the text is basically a 1D signal (symbols to be recognized are organized in lines of fixed height, in which they extend from left to right or vice versa), musical notation can additionally be stacked arbitrarily also on the vertical axis, thus becoming a 2D signal. This superposition property would exponentially increase the number of symbols to be recognized, if approached the usual way (which is intractable from a computational as well as from a classification point of view). It also makes segmentation very hard and does not imply a natural ordering of the symbols as for example in the SVHN dataset \cite{Netzer2011}.

In this paper, we present the \emph{DeepScores} dataset with the following contributions: a) a curated dataset of a collection of hundreds of thousands of musical scores, containing tens of millions of objects to construct a high quality dataset of written music; b) available ground truth for the tasks of object detection, semantic segmentation, and classification; c) comprehensive comparisons with other computer vision datasets (see Section \ref{sec_comparison}) and a quantitative and qualitative analysis of \emph{DeepScores} (see Section \ref{sec_statistics}); d) computation of an object classification baseline and a qualitative assessment of current off-the-shelf detection methods along with reasoning why detection needs new approaches on \emph{DeepScores} (see Section \ref{sec_baselines}); e) proposals on how to facilitate next generation computer vision research using \emph{DeepScores} (see Section \ref{sec_conclusions}). The data, a recommended evaluation scheme  and accompanying TensorFlow \cite{Tensorflow} code are freely available\footnote{\url{https://tuggeluk.github.io/deepscores/}}.

\begin{figure}[t]
\centering
\includegraphics[width=0.5\textwidth]{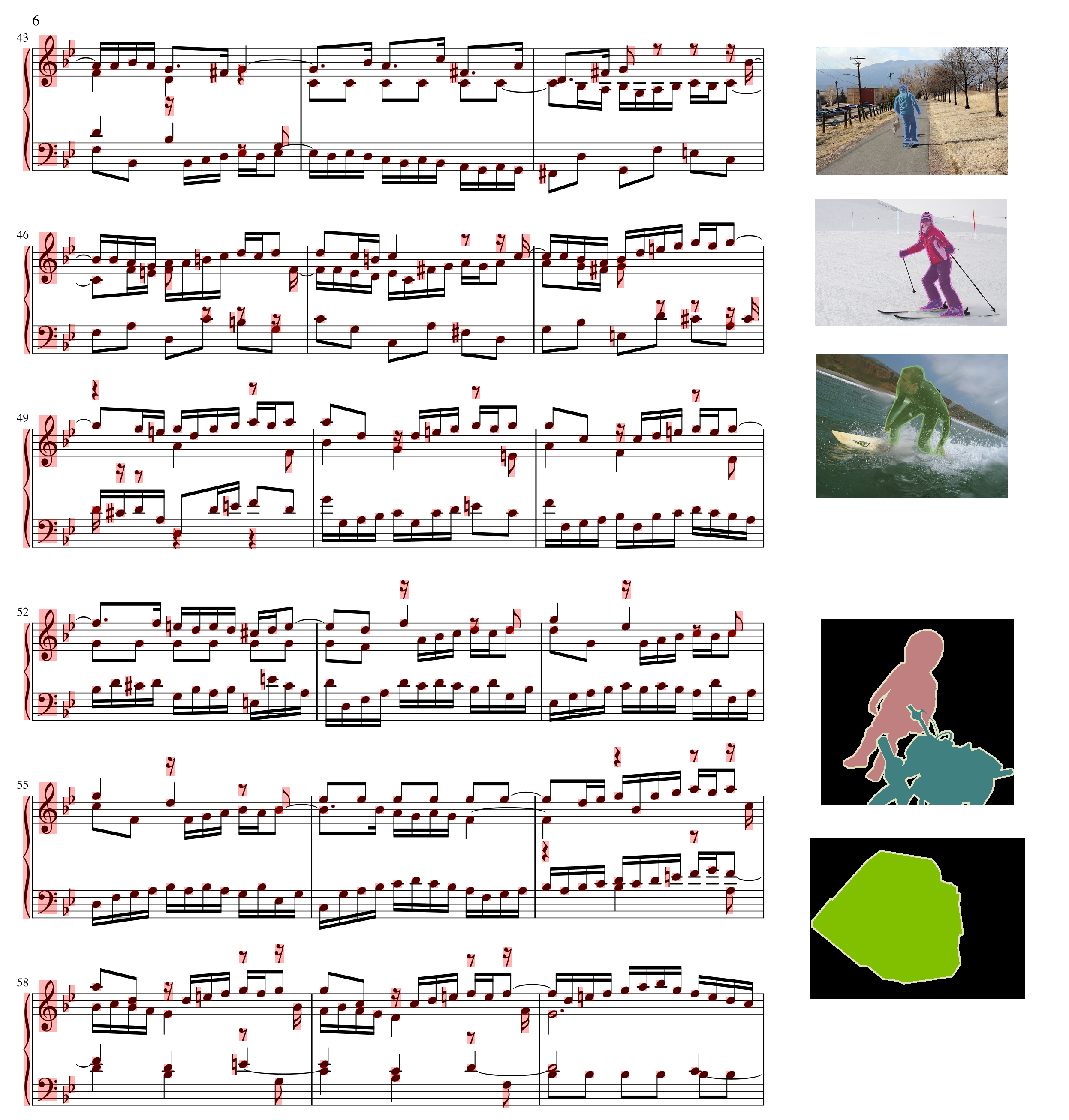}
\caption{A typical image and ground truth from the \emph{DeepScores} dataset (left), next to examples from the MS-COCO (3 images, top right) and PASCAL VOC (2, bottom right) datasets. Even though the music page is rendered at a much higher resolution, the objects are still smaller; the size ratio between the images is truthful despite all images being downscaled.}
\label{fi:1}
\end{figure}

\section{\emph{DeepScores} in the context of other datasets}
\label{sec_comparison}

\begin{figure}[t]
\begin{center}
\subfloat[Snippet of an input image.]      
{
\includegraphics[width=.6\linewidth]{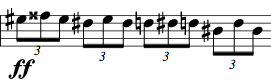}
\label{fi:2:a}
}\\
\subfloat[Boundig boxes rendered over single objects from snippet \ref{fi:2:a} for object detection.]   {
\includegraphics[width=.6\linewidth]{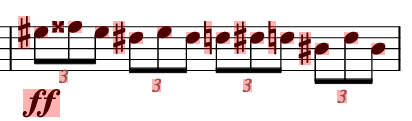}
\label{fi:2:b}
}\\
\subfloat[Color-based pixel level labels (the differences are hard to recognize visually, but there is a distinct color per symbol class) for semantic segmentation.]     
{
\includegraphics[width=.6\linewidth]{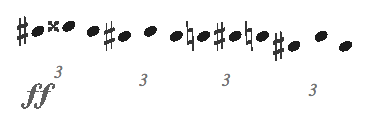}
\label{fi:2:c}
}\\
\subfloat[Patches centered around specific symbols (in this case: \texttt{gClef}) for object classification.]      
{
\includegraphics[width=.4\linewidth]{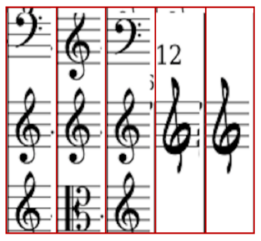}
\label{fi:2:d}
}
\end{center}  
\caption{Examples for the different flavors of ground truth available in \emph{DeepScores}.}
\label{fi:2}  
\end{figure}

\emph{DeepScores} is a high quality dataset consisting of pages of written music, rendered at $400$ dots per inch (dpi). It has $300,000$ full pages as images, containing tens of millions of objects, separated into $123$ classes (cp. Figure \ref{fi:1}). The aim of the dataset is to facilitate general research on small object recognition, with direct applicability to the recognition of musical symbols. We provide the dataset with ground truth for the following tasks: object classification, semantic segmentation, and object detection (cp. Figure \ref{fi:2}).

\textbf{Object classification} in the context of computer vision is the procedure of labeling an image with a single label. Its recent history is closely linked to the success of deep convolutional learning models \cite{Fukushima1980, LeCun1990}, leading to superhuman performance \cite{ciresan2012cvpr} and subsequent ImageNet object classification breakthroughs \cite{Krizhevsky2012}. Shortly afterwards, similar systems achieved human-level accuracy also on ImageNet \cite{Srivastava2015, Szegedy2015, He2016}. Generally speaking, the ImageNet dataset \cite{Deng2009} was a key ingredient to the success of image classification algorithms.

In \emph{DeepScores}, we provide data for the classification task even though classifying musical symbols in isolation is not a challenging problem compared to classifying ImageNet images. But providing the dataset for classification, in addition to a neural network implementation that achieves high accuracy (see Section \ref{sec_baselines}), might help to address the other two tasks. In fact, the first step in many computer vision models is to use a deep convolutional neural network pre-trained on ImageNet, and alter it for the task of image segmentation or image detection \cite{Long2015, Ren2014}. We expect that the same technique can be used when it comes to OMR and detecting very small objects.

\textbf{Semantic segmentation} is the task of labeling each pixel of the image with one of the possible classes. State-of-the-art models are typically based on fully convolutional architectures \cite{Ciresan2012isbi, Long2015}. The task is arguably a significantly more difficult problem than image classification, with the recent success being largely attributed to the release of high quality datasets like PASCAL VOC \cite{Everingham2010} and MS-COCO \cite{Lin2014}. 

In \emph{DeepScores}, we provide ground truth for each pixel in all the images, having roughly $10^{12}$ labeled pixels in the dataset.

\textbf{Object detection} is the by far most interesting and challenging task: to classify all the objects in the image, and at the same time to find their precise position. State-of-the-art algorithms are pipeline convolutional models, typically having combined cost functions for detection and classification \cite{Ren2014, Liu2016, Redmon2016}. Similar to the case of semantic segmentation above, the PASCAL VOC and especially MS-COCO datasets have played an important part on the recent success of object detection algorithms.

In \emph{DeepScores}, we provide bounding boxes and labels for each of the musical symbols in the dataset. With around $80$ million objects, this makes our dataset the largest one released so far, and highly challenging. More on the challenges of \emph{DeepScores} is provided in section \ref{sec_baselines}.

\subsection{Comparisons with computer vision datasets}
\label{comparison}

Compared with some of the most used datasets in the field of computer vision, \emph{DeepScores} has by far the largest number of objects, as well as the highest resolution. In particular, images of \emph{DeepScores} typically have a resolution of $1,894$ x $2,668$ pixels, which is at least four times higher than the resolutions of datasets we compare with. Table \ref{num_of_classes} contains quantitative comparisons of \emph{DeepScores} with other datasets, while the following paragraphs bring in also qualitative aspects.

\textbf{SVHN}, the street view house numbers dataset \cite{Netzer2011}, contains $600,000$ labeled digits cropped from street view images. Compared to \emph{DeepScores}, the number of objects in SVHN is two orders of magnitude lower, and the number of objects per image is two to three orders of magnitude lower.

\textbf{ImageNet} contains a large number of images and (as a competition) different tracks (classification, detection and segmentation) that together have proven to be a solid foundation for many computer vision projects. However, the objects in ImageNet are quite large, while the number of objects per image is very small. Unlike ImageNet, \emph{DeepScores} tries to address this issue by going to the other extreme, providing a very large number of very small objects on images having significantly higher resolution than all the other mentioned datasets.

\textbf{PASCAL VOC} is a dataset which has been assembled mostly for the tasks of detection and segmentation. Compared to ImageNet, the dataset has slightly more objects per image, but the number of images is comparatively small: our dataset is one order of magnitude bigger in the number of images, and three orders of magnitude bigger in the number of objects.

\textbf{MS-COCO} is a large upgrade over PASCAL VOC on both the number of images and number of objects per image. With more than $300$k images containing more than $3$ million objects, the dataset is very useful for various tasks in computer vision. However, like ImageNet, the number of objects per image is still more than one order of magnitude lower than in our dataset, while the objects are relatively large. 

\begin{table}[t]
\centering
\begin{tabular}{ |p{2.15cm}||p{1.0cm}|p{1.0cm}|p{1.cm}|p{1.cm}| }
 \hline
 \bf{Dataset} & \bf{\#classes} & \bf{\#images} & \bf{\#objects} & \bf{\#pixels}\\
 \hline
 \hline
 SUN & \textbf{397} & 17k & 17k & 6b\\
 \hline
 PASCAL VOC & 21 & 10k & 30k & 2.5b\\
 \hline
 MS COCO & 91 & 330k &3.5m & 100b\\
 \hline
 ImageNet & 200 & \textbf{500k} & 600k & 125b\\
 \hline
 SVHN & 10 & 200k & 630k & 4b\\
 \hline
 \emph{DeepScores} & 123 & 300k & \textbf{80m} & \textbf{1.5t}\\
 \hline
\end{tabular}
\caption{Information about the number of classes, images and objects for some of the most common used datasets in computer vision. The number of pixels is estimated due to most datasets not having fixed image sizes. We use the SUN 2012 object detection specifications for SUN, and the statistics of the ILSVRC 2014 \cite{Russakovsky2015} detection task for ImageNet.}
\label{num_of_classes}
\end{table}

\subsection{Comparisons with OMR datasets}

A number of OMR datasets have been released in the past with a specific focus on the computer music community. \emph{DeepScores} will be of use both for general computer vision as well as to the OMR community (compare Section \ref{sec_baselines}).

\paragraph{Handwritten scores}\mbox{}

The Handwritten Online Musical Symbols dataset \textbf{HOMUS} \cite{Zaragoza2014} is a reference corpus with around $15,000$ samples for research on the recognition of online handwritten music notation. For each sample, the individual strokes that the musician wrote on a Samsung tablet using a stylus were recorded and can be used in online and offline scenarios.

The \textbf{CVC-MUSCIMA} database \cite{Fornes2012} contains handwritten music images, which have been specifically designed for writer identification and staff removal tasks. The database contains $1,000$ music sheets written by $50$ different musicians with characteristic handwriting styles. 

\textbf{MUSCIMA++} \cite{Hajic2017} is a dataset of handwritten music for musical symbol detection that is based on the MUSCIMA dataset. It contains $91,255$ written symbols, consisting of both notation primitives and higher-level notation objects, such as key signatures or time signatures. There are $23,352$ notes in the dataset, of which $21,356$ have a full notehead, $1,648$ have an empty notehead, and $348$ are grace notes.  

The \textbf{Capitan Collection} \cite{Zaragoza2016} is a corpus collected via an electronic pen while tracing isolated music symbols from early manuscripts. The dataset contains information on both the sequence followed by the pen (capitan stroke) as well as the patch of the source under the tracing itself (capitan score). In total, the dataset contains $10,230$ samples unevenly spread over $30$ classes.

\paragraph{Print quality scores}\mbox{}

The \textbf{MuseScore Monophonic MusicXML Dataset} \cite{vanderWel17} is one of the largest OMR dataset to date, consisting of $17,000$ monophonic scores. While the dataset has high quality images, it doesn't resemble real-world musical scores which are not monophonic and thus have many lines per image. 

Further OMR datasets of printed scores are reviewed by the \textbf{OMR-Datasets} project\footnote{See \url{https://apacha.github.io/OMR-Datasets/}.}. \emph{DeepScores} is by far larger than any of these or the above-mentioned datasets, containing more images and musical symbols than all the other datasets combined. In addition, \emph{DeepScores} contains only real-world scores (i.e., symbols in context as they appear in real written music), while most other datasets are either synthetic or reduced (containing only symbols in isolation or just a line per image). The sheer scale of \emph{DeepScores} makes it highly usable for modern deep learning algorithms. While convolutional neural networks have been used before for OMR \cite{vanderWel17}, \emph{DeepScores} for the first time enables the training of very large and deep models. 

\section{The \emph{DeepScores} dataset}
\label{sec_statistics}

\subsection{Quantitative properties}
\emph{DeepScores} contains around $300,000$ pages of digitally rendered music scores (see Sections \ref{sec:construction} and \ref{sec:challenges} for a justification of synthetic data) and has ground truth for $123$ different symbol classes. The number of labeled music symbol instances is roughly $80$ million ($4$-$5$ orders of magnitude higher than in the other music datasets; when speaking of symbols, we mean labeled musical symbols that are to be recognized as objects in the task at hand). The number of symbols on one page can vary from as low as $4$ to as high as $7,664$ symbols. On average, a sheet (i.e., an image) contains around $243$ symbols. Table \ref{statistics} gives the mean, standard deviation, median, maximum and minimum number of symbols per page in the ``symbols per sheet'' column.

\begin{table}[t] 
\centering
\begin{tabular}{ |l||r|r| }
 \hline
 \bf{Statistic} & \bf{Symbols per sheet} & \bf{Symbols per class} \\
 \hline
 \hline
 Mean & $243$ & $650$k\\
 \hline
 Std. dev. & $203$ & $4$m\\
 \hline
 Maximum & $7'664$ & $44$m\\
 \hline
 Minimum & $4$ & $18$\\
 \hline
 Median & $212$ & $20$k\\
 \hline
\end{tabular}
\caption{Statistical measures for the occurrence of symbols per musical sheet and per class (rounded).}
\label{statistics}
\end{table}

Another interesting aspect of \emph{DeepScores} is the class distribution. Obviously, some classes contain more symbols than other classes (see also Table \ref{statistics}, column 3). It can be seen that the average number of elements per class is $650$k but the standard deviation is $4$m, illustrating that the distribution of symbols per class is very unbalanced.

\subsection{Flavors of ground truth}

In order for \emph{DeepScores} to be useful for as many applications as possible, we offer ground truth for three different tasks. For object classification, there are up to $3,000$ labeled image patches per class, i.e. we do not provide each of the $80$m symbols as a single patch for classification purposes. Instead, we constrain the dataset for this simpler task to a random subset of reasonable size (see Section \ref{sec_baselines}). The patches have size $220\times 120$ and contain the full original context of the symbol (i.e., they are cropped out of real world musical scores). Each patch is centered around the symbol's bounding box (see Figure \ref{fi:2:d}). 

For object detection, an accompanying XML file for each image in \emph{DeepScores} holds an \texttt{object} node for each symbol instance present on the page. It contains its class and bounding box coordinates, visualized in Figure \ref{fi:2:b}. 

For semantic segmentation, there is an accompanying PNG file for each image. This PNG has identical size as the initial image, but each pixel has been recolored to represent the symbol class it is part of. As in Figure \ref{fi:2:c}, the background is white, with the published images using grayscale colors from $0$ to $123$ for ease of use in the softmax layer of potential models.

\subsection{Dataset construction}
\label{sec:construction}

\emph{DeepScores} is constructed by synthesizing sheet music from a large collection of written music in a digital format: crowd-sourced MusicXML files publicly available from MuseScore\footnote{See \url{https://musescore.com}.} and used by permission. The rendering of MusicXML and the generation of accompanying ground truth is one of the main contributions of this work. Going from online MusicXML archives to a curated dataset is non-trivial due to extensive musical know-how and non-available custom software components necessary to create examples containing symbols and \emph{corresponding} object locations. It involves a) code to be injected in the LilyPond\footnote{See \url{http://lilypond.org/}.} SVG backend such that the printed SVG paths contain additional meta data for each individual symbol; b) software that maps each found path to one of the predefined object classes and renders colored PNG files correctly (i.e., crisp edges, exact localization etc) as well as XML descriptions; and c) software that constructs the final ground truth out of generated meta data. All steps have been aligned with musicians to guarantee fitness for the OMR task.

\begin{figure}[t]
\centering
\includegraphics[width=0.4\textwidth]{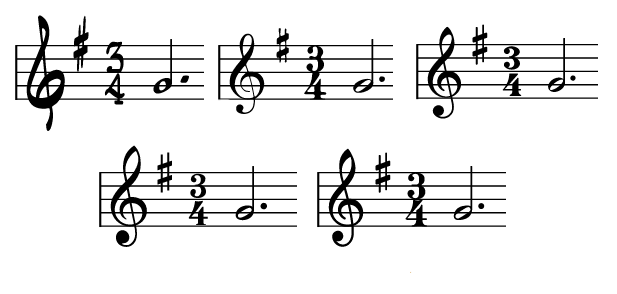}
\caption{The same patch of a musical sheet, rendered using five different fonts.}
\label{fi:7}
\end{figure}

To achieve a realistic variety in the data even though all images are digitally rendered and therefore have perfect image quality, five different music fonts have been used for rendering (see Figure \ref{fi:7}). The challenge of the dataset however is not in the variety of the presentation of the different symbol instances, as is the case with traditional object detection datasets (see Section \ref{sec:challenges}).

A key feature of a dataset is the definition of the classes to be included. Due to their compositional nature, there are many ways to define classes of music symbols: is it for example a ``c'' note with duration $8$ (\texttt{cNote8th}) or is it a black notehead (\texttt{noteheadBlack}) and a flag (\texttt{flag8thUp} or \texttt{flag8thDown})? Adding to this complexity, there is a huge number of special and thus infrequent symbols in music notation. The selected set is the result of many discussions with music experts and contains the most important symbols. We decided to use atomic symbol parts as classes which makes it possible to define composite symbols in an application-dependent way. As a result of these discussions we also decided to focus on fixed-shape symbols and have left out stems, barlines, staff and ledger lines. 

\section{Experiments and impact}
\label{sec_baselines}

\subsection{Unique challenges}
\label{sec:challenges}

\begin{figure}[h]

\begin{center}
\subfloat[\texttt{accidentalSharp}]      {
\includegraphics[width=.2\linewidth]{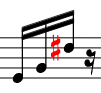}
}\quad
\subfloat[\texttt{keySharp}]      {
\includegraphics[width=.2\linewidth]{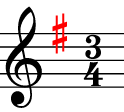}
}\\
\subfloat[\texttt{augmentationDot}]      {
\includegraphics[width=.25\linewidth]{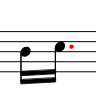}
}\quad
\subfloat[\texttt{articStaccatoAbove}]      {
\includegraphics[width=.25\linewidth]{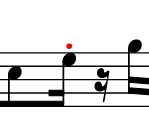}
}
\end{center}
\caption{Examples of the importance of context for classifying musical symbols: in both rows, the class of otherwise similar looking objects changes depending on the surrounding objects.}
\label{fi:6}
\end{figure}

One of the key challenges \emph{DeepScores} poses upon modeling approaches is the sheer amount of objects on a single image. There are two additional properties of music notation imposing challenges. First, there is a big variability in object size ranging from less than hundred to many thousands of pixels in area. Second, context matters in music notation: two objects having the same appearance can belong to a different class depending on the local surroundings (see Figure \ref{fi:6}). To our knowledge there is no other freely available large scale dataset that shares this trait. 


Moreover, datasets like ImageNet are close to being perfectly balanced, with the number of images and objects per class being a constant. This clearly isn't the case with the \emph{DeepScores} dataset, where the most common class contains more than half of the symbols in the dataset, and the top $10$\% of classes contain more than $85$\% of the symbols in the entire dataset. This extremely skewed distribution resembles many real-world cases for example in anomaly detection and industrial quality control.

\subsection{Analysis of off-the-shelf deep learning models}

Merely classifying the musical symbols in \emph{DeepScores} is expected to be relatively simple: all symbols have very clear black and white borders, their shape has limited variability and they are rendered at a very high resolution. We fitted a simple residual CNN \cite{He2016} with $25$ convolutional layers and about $8$ million trainable parameters. Using the Adam optimizer with the hyperparameters proposed by the authors \cite{adam}, we reached a macro average accuracy of over $0.98$ in about ten epochs. This shows that CNNs are able to deal with labels that not only depend on an object, but also its surroundings.

Detection, however, is more challenging. This is due to the sheer amount of small objects present on each image, which stands in stark contrast to the low number of prominent objects present in natural images. Since the leading detection systems SSD, YOLO and Faster R-CNN have been developed with natural images in mind, \emph{DeepScores} is a tough challenge for them. Evaluating the input on a fixed $S\times S$ grid makes YOLO inherently unfit to deal with clusters of small symbols: \textit{"YOLO imposes strong spatial constraints on bounding box predictions since each grid cell only predicts two boxes and can only have one class. This spatial constraint limits the number of nearby objects that our model can predict.
Our model struggles with small objects that appear in
groups, such as flocks of birds"} \cite{Redmon2016}. SSD uses features from six of the top layers, but it still struggles with small objects as visible in Figure 4 of the original publication \cite{Liu2016}. 

Therefore, we ran experiments only with Faster-RCNN, using smaller anchors as in the published configuration to adapt to our task. The system was unable to find any symbols at all. It is unclear whether fine tuning the region proposal network and anchor setup will lead  to a good performance on \emph{DeepScores}. Instead, a novel detection method currently under development and based on fully convolutional neural networks \cite{Long2015} shows promising preliminary results on \emph{DeepScores}, validating our intuition that the dataset is well-suited for the development of new methods focused on many tiny objects.

\subsection{Expected impact}

Both observations---easy classification but challenging detection---lie at the heart of what we think makes \emph{DeepScores} very useful: on the one hand, it offers the challenging scenario of many tiny objects that cannot be approached using current datasets. On the other hand, \emph{DeepScores} is probably the easiest scenario of that kind, because classifying single musical objects is relatively easy and the dataset contains a vast amount of training data. \emph{DeepScores} thus is a prime candidate to develop next generation document recognition and computer vision methods that scale to many tiny objects on large images: many real-world problems deal with high-resolution images, with images containing hundreds of objects and with images containing very small objects in them. This might be OMR itself, automated driving and other robotics use cases, medical applications with full-resolution imaging techniques as data sources, vision-based industrial quality control, or surveillance tasks e.g. in sports arenas and other public places. 

Finally, \emph{DeepScores} will be a valuable source for pre-training models: 
transfer learning has been one of the most important ingredients in the advancement of computer vision. The first step in many computer vision models \cite{Long2015,Ren2014} is to use a deep convolutional neural network pre-trained on ImageNet, and alter it for the task of image segmentation or object detection, or use it on considerably smaller, task-dependent final training sets. 

\section{Conclusions}
\label{sec_conclusions}

We have presented the conception and creation of \emph{DeepScores} - one of the largest publicly and freely available datasets for OMR and computer vision applications in general in terms of image size and number of contained objects. Compared to other well-known datasets, \emph{DeepScores} has large images (more than four times larger than the average) containing many (one to two orders of magnitude more) very small (down to a few pixels, but varying by several orders of magnitude) objects that change their class belonging depending on the visual context. The dataset is made up of sheets of written music, synthesized from the largest public corpus of MusicXML. It comprises ground truth for the tasks of object classification, semantic segmentation and object detection. 

We have argued that the unique properties of \emph{DeepScores} make the dataset suitable for use in the development of general next generation computer vision methods that are able to work on large images with tiny objects. This ability is crucial for real-world applications like robotics, automated driving, medical image analysis, industrial quality control or surveillance, besides OMR. We have motivated that object classification is relatively easy on \emph{DeepScores}, making it therefore the potentially cheapest way to work on a challenging detection task. We thus expect impact on general object detection algorithms. 

One weakness of the \emph{DeepScores} dataset is that all the data is currently digitally rendered. Linear models (or piecewise linear models like neural networks) have been shown to not generalize well when the distribution of the real-world data is far from the distribution of the dataset the model has been trained on \cite{Torralba2011, Szegedy2013}. Our experiments show that networks trained on \emph{DeepScores} do generalize to high quality scans, but processing lower quality images remains a challenge. To address this issue, we currently construct training data that consist of flatbed-scans and photos of low-res prints. Many colleagues mentioned that ground truth for non-fixed shape symbols (e.g. slurs, beams) is of high importance to them, therefore are we working \emph{on} an updated version of \emph{DeepScores} that carries this information.

Future work \emph{with} the dataset will -- besides the general impact predicted above -- directly impact OMR: the full potential of deep neural networks is still to be realized on musical scores. 

\section*{Acknowledgements} This work is financially supported by CTI grant 17963.1 PFES-ES ``DeepScore''. The authors are grateful for the support of Herv\'{e} Bitteur of Audiveris, the permission to use MuseScore data, and the collaboration with ScorePad AG.


\begin{thebibliography}{1}

\bibitem{Russakovsky2015}
O. Russakovsky, J. Deng, H. Su, J. Krause, S. Satheesh, S. Ma, Z. Huang, A. Karpathy, A. Khosla, A. C. Berg and L. Fe-Fei, \emph{ImageNet Large Scale Visual Recognition Challenge}, International Journal of Computer Vision, 2015.

\bibitem{mori1999}
M. Shunji, N. Hirobumi, Y. Su and J. Hiromitsu, \emph{Optical character recognition}, John Wiley \& Sons, Inc., 1999.

\bibitem{Lee16}
C. Y. Lee and S. Osindero, \emph{Recursive Recurrent Nets with Attention Modeling for {OCR} in the Wild}, Computer Vision and Pattern Recognition, 2016.

\bibitem{vanderWel17}
E. van der Wel and K. Ullrich, \emph{Optical Music Recognition with Convolutional Sequence to-Sequence Models}, http://arxiv.org/abs/1707.04877, 2017.

\bibitem{Rebelo2012}
A. Rebelo, I. Fujinaga, F. Paszkiewicz, A. R. S. Marcal, C. Guedes and J. S. Cardaso, \emph{Optical music recognition: state-of-the-art and open issues}, International Journal of Multimedia Information Retrieval, 2012.

\bibitem{ciresan2012cvpr}
D. C. Ciresan, U. Meier and J. Schmidhuber, \emph{Multi-Column Deep Neural Networks for Image Classification}, Computer Vision and Pattern Recognition, 2012.

\bibitem{Ciresan2011}
D. C. Ciresan, U. Meier, J. Masci, L. M. Gambardella and J. Schmidhuber, \emph{High-Performance Neural Networks for Visual Object Classification}, arXiv:1102.0183v1  [cs.AI], 2011.

\bibitem{Fukushima1980}
K. Fukushima, \emph{Neocognitron: A self-organizing neural network for a mechanism of pattern recognition unaffected by shift in position}, Biological Cybernetics, 1980.

\bibitem{LeCun1990}
Y. LeCun, B. E. Boser, J. S. Denker, D. Henderson, R. E. Howard, W. E. Hubbard and L. D. Jackel, \emph{Handwritten digit recognition with a back-propagation network.}, NIPS, 1990.

\bibitem{gpu2004}
K-S. Oh and K. Jung, \emph{{GPU} implementation of neural networks}, Pattern Recognition, 2004.

\bibitem{Srivastava2015}
R. K. Srivastava, K. Greff and J. Schmidhuber, \emph{Highway networks}, arXiv preprint arXiv:1505.00387, 2015.

\bibitem{adam}
D. Kingma and J. Ba, \emph{Adam: A method for stochastic optimization}, International Conference on Learning Representations, 2015.

\bibitem{Ciresan2012isbi}
D. C. Ciresan, A. Giusti, L. M. Gambardella and J. Schmidhuber, \emph{Neural Networks for Segmenting Neuronal Structures in {EM} Stacks}, ISBI Segmentation Challenge Competition: Abstracts, 2012.

\bibitem{Raina2009}
R. Raina, A. Madhavan and A. Y. Ng, \emph{Large-scale deep unsupervised learning using graphics processors}, ICML, 2009.
  
\bibitem{Deng2009}
J. Deng, W. Dong, R. Socher, J. L. Li and L. Fei-Fei, \emph{Imagenet: A large-scale hierarchical image database.}, Computer Vision and Pattern Recognition, 2009.  
  
\bibitem{Everingham2010}
M. Everingham, L. Van Gool, C. K. Williams, J. Winn and A. Zisserman, \emph{The pascal visual object classes (voc) challenge.}, International journal of computer vision, 2010.    
  
\bibitem{Xiao2010}
J. Xiao, J. Hays, K. A. Ehinger, A. Oliva and A. Torralba, \emph{Sun database: Large-scale scene recognition from abbey to zoo.}, Computer Vision and Pattern Recognition, 2010.   
  
\bibitem{Netzer2011}
Y. Netzer, T. Wang, A. Coates, A. Bissacco, B. Wu and A. Y. Ng, \emph{Reading digits in natural images with unsupervised feature learning.}, NIPS workshop, 2011.    
  
\bibitem{Lin2014}
T. Y. Lin, M. Maire, S. Belongie, J. Hays, P. Perona, D. Ramanan and C. L. Zitnick, \emph{Microsoft coco: Common objects in context.}, European Conference in Computer Vision, 2014.    
 
\bibitem{Ren2014}
S. Ren, K. He, R. Girshick and J. Sun, \emph{Faster R-CNN: Towards real-time object detection with region proposal networks.}, NIPS, 2014.   
  
\bibitem{Liu2016}
W. Liu, D. Anguelov, D. Erhan, C. Szegedy, S. Reed, C. Y. Fu and A. C. Berg, \emph{Ssd: Single shot multibox detector.}, European Conference in Computer Vision, 2016.      
  
\bibitem{Redmon2016}
J. Redmon, S. Divvala, R. Girshick and A. Farhadi, \emph{You only look once: Unified, real-time object detection.}, ICCV, 2016.     
  
\bibitem{Goodfellow2013}
I. J. Goodfellow, Y. Bulatov, J. Ibarz, S. Arnoud and V. Shet, \emph{Multi-digit number recognition from street view imagery using deep convolutional neural networks.}, arXiv preprint arXiv:1312.6082., 2013.   
  
\bibitem{Krizhevsky2012}
A. Krizhevsky, I. Sutskever and G. E. Hinton, \emph{Imagenet classification with deep convolutional neural networks.}, NIPS, 2012.     
  
\bibitem{Szegedy2015}
C. Szegedy, W. Liu, Y. Jia, P. Sermanet, S. Reed, D. Anguelov, V. Vanhoucke and A. Rabinovich, \emph{Going deeper with convolutions.}, Computer Vision and Pattern Recognition, 2015.   
  
\bibitem{He2016}
K. He, X. Zhang, S. Ren and J. Sun, \emph{Deep residual learning for image recognition.}, Computer Vision and Pattern Recognition, 2016.      
  
\bibitem{Long2015}
J. Long, E. Shelhamer and T. Darrell, \emph{Fully convolutional networks for semantic segmentation.}, Computer Vision and Pattern Recognition, 2015.    

\bibitem{Zaragoza2014}
J. Calvo-Zaragoza and J. Oncina, \emph{Recognition of Pen-Based Music Notation: The HOMUS Dataset.}, ICPR, 2014.   

\bibitem{Fornes2012}
A. Fornes, A. Dutta, A. Gordo and J. Llados, \emph{A Ground-truth of Handwritten Music Score Images for Writer Identification and Staff Removal.}, International Journal on Document Analysis and Recognition, 2012.   

\bibitem{Hajic2017}
J. Hajic and P. Pecina, \emph{The MUSCIMA++ Dataset for Handwritten Optical Music Recognition.}, ICDAR 2017

\bibitem{Zaragoza2016}
J. Calvo-Zaragoza, D. Rizo and J. M. Inesta, \emph{Two (note) heads are better than one: pen-based multimodal interaction with music scores.}, International Society of Music Information Retrieval conference, 2016.   

\bibitem{jorge2017}
Jorge Calvo{-}Zaragoza,Jose J. Valero{-}Mas and
Antonio Pertusa, \emph{End-to-End Optical Music Recognition Using Neural Networks}, ISMRIR 2017

\bibitem{Torralba2011}
A. Torralba and A. Efros, \emph{Unbiased look at dataset bias}, Computer Vision and Pattern Recognition, 2011

\bibitem{Szegedy2013}
C. Szegedy, W. Zaremba, I. Sutskever, J. Bruna, R. Erhan, I. Goodfellow and R. Fergus, \emph{Intriguing properties of neural networks.}, arXiv:1312.6199., 2013. 

\bibitem{Tensorflow}
Martin Abadi et al., \emph{Tensorflow: Large-scale machine learning on heterogeneous distributed systems.}, arXiv preprint arXiv:1603.04467, 2016.


\end{thebibliography}
\end{document}